\documentclass[final]{cvpr}

\usepackage{times}
\usepackage{epsfig}
\usepackage{graphicx}
\usepackage{amsmath}
\usepackage{amssymb}
\usepackage{bm}
\usepackage{algorithm}
\usepackage{algorithmic}
\usepackage{multirow}
\usepackage{eqparbox}
\usepackage{color}
\usepackage{float}
\usepackage{makecell}

\usepackage[pagebackref=true,breaklinks=true,colorlinks,bookmarks=false]{hyperref}

\begin{document}

\title{Self-Point-Flow: Self-Supervised Scene Flow Estimation from Point Clouds with Optimal Transport and Random Walk}

\author{Ruibo Li$^{1,2}$, ~~ Guosheng Lin$^{1,2}$\thanks{Corresponding author: G. Lin. (e-mail:{ $\tt gslin@ntu.edu.sg$ })}, ~~Lihua Xie$^3$\\
	$^{1}$S-Lab, Nanyang Technological University, Singapore\\
	$^{2}$School of Computer Science and Engineering, Nanyang Technological University, Singapore \\
	$^{3}$School of Electrical and Electronic Engineering, Nanyang Technological University, Singapore\\
	E-mail: { $\tt ruibo001@e.ntu.edu.sg$ }, { $\tt \{gslin, elhxie\}@ntu.edu.sg$ }
}

\maketitle

\begin{abstract}
Due to the scarcity of annotated scene flow data, self-supervised scene flow learning in point clouds has attracted increasing attention. In the self-supervised manner, establishing correspondences between two point clouds to approximate scene flow is an effective approach. Previous methods often obtain correspondences by applying point-wise matching that only takes  the distance on 3D point coordinates into account, introducing two critical issues: (1) it overlooks other discriminative measures, such as color and surface normal, which often bring fruitful clues for accurate matching; and (2) it often generates sub-par performance, as the matching is operated in an unconstrained situation, where multiple points can be ended up with the same corresponding point. To address the issues, we formulate this matching task as an optimal transport problem. The output optimal assignment matrix can be utilized to guide the generation of pseudo ground truth. In this optimal transport, we design the transport cost by considering multiple descriptors and encourage one-to-one matching by mass equality constraints. Also, constructing a graph on the points, a random walk module is introduced to encourage the local consistency of the pseudo labels. Comprehensive experiments on FlyingThings3D and KITTI show that our method achieves state-of-the-art performance among self-supervised learning methods. Our self-supervised method even performs on par with some supervised learning approaches, although we do not need any ground truth flow for training.

\end{abstract}

\section{Introduction}

Scene flow estimation aims to obtain a  3D vector field of points in dynamic scenes, and describes the motion state of each point. 
Recently, with the popularity of 3D sensors and the great success of deep learning in 3D point cloud tasks, directly estimating the scene flow from point clouds by deep neural networks (DNNs) is an active research topic.
DNNs are data-driven, and the supervised training of DNNs requires a large amount of training data with ground truth labels. 
However, for the scene flow estimation task, no sensor can capture optical flow ground truth in complex environments~\cite{menze2015object}, which makes real-world scene flow ground truth hard to obtain.
Due to the scarcity of the ground truth data, recent deep learning based point cloud scene flow estimation methods~\cite{liu2019flownet3d,gu2019hplflownet,wu2020pointpwc,puy2020flot} turn to synthetic data, $e.g.$ the FlyingThings3D dataset~\cite{mayer2016large}, for supervised training. 
However, the domain gap between synthetic data and realistic data is much likely to make the trained models perform poorly in real-world scenes.

\begin{figure}[tb]
	\centering
	\includegraphics[height=5.0cm]{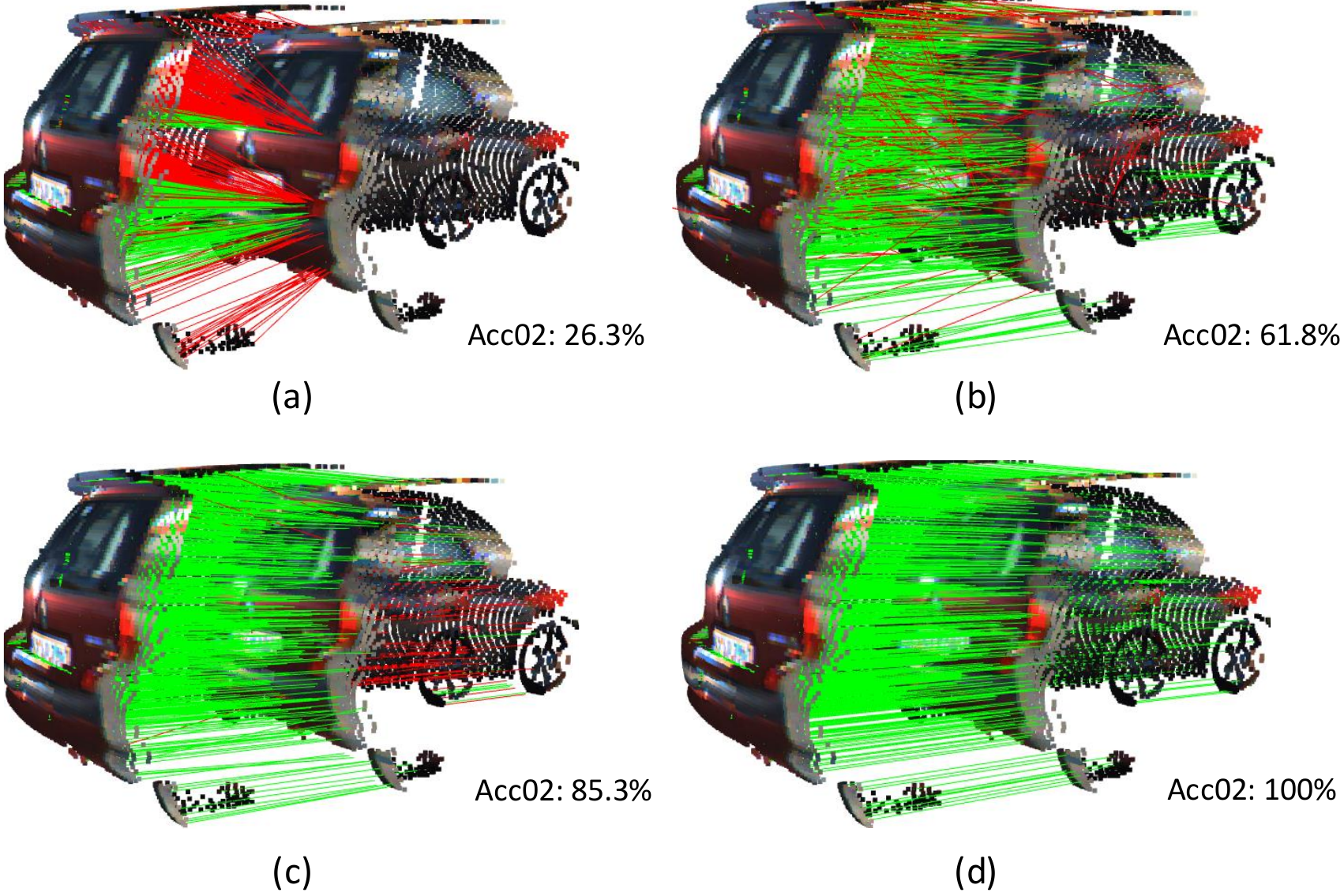}
	\caption{ Pseudo ground truth between two point clouds produced by different methods.
		The green line represents the correct pseudo ground truth whose relative error is less than 10\%.
		The red line represents the wrong pseudo ground truth.
		(a) pseudo ground truth produced by nearest neighbor search with only 3D point coordinate as measure; 
		(b) pseudo ground truth produced by our optimal transport module; 
		(c) pseudo ground truth produced by our optimal transport module and then refined by our random walk module; 
		(d) ground truth.
	}
	\label{introduction}	
\end{figure}

To circumvent the dependence on expensive ground truth data, we target self-supervised scene flow estimation from point clouds.
Mittal et al.~\cite{mittal2020just} and Wu et al.~\cite{wu2020pointpwc} make the first attempt.
These methods search for the closest point in the other point cloud as the corresponding point and use the coordinate difference between each correspondence to approximate the ground truth scene flow.
Although achieving promising performance, two issues exist in these methods:
(1) searching for correspondences relies only on 3D point coordinates but ignores other measures, such as color and surface normal, which often bring fruitful clues for accurate matching;
and (2) the unconstrained search may lead to a degenerated solution, where multiple points match with the same point in the other point cloud, $i.e.$ a many-to-one problem.
An example of nearest neighbor search is shown in Fig.~\ref{introduction}(a).

In this paper, we assume that an object's geometric structure and appearance remain unchanged as it moves and the correct corresponding points could be found in its neighborhood.
Thus, when searching for point correspondences, we adopt 3D point coordinate, surface normal, and color as measures and encourage each point to be matched with a unique one in the next frame, $i.e.$ one-to-one matching.
Naturally, the searching  problem can be formulated as an optimal transportation~\cite{peyre2019computational}, where the transport cost is defined on the three measures, the mass equality constraints are built to encourage one-to-one matching, and the produced optimal assignment matrix indicates the optimal correspondences between the two point clouds.
Removing some invalid correspondences with far distance, the coordinate differences between valid correspondences can be treated as the pseudo ground truth flow vectors for training.

Neighboring points in an object often share a similar movement pattern.
However, the optimal transport module generates pseudo labels by point-wise matching without considering the local relations among neighboring points, resulting in conflicting pseudo labels in each local region, as shown in Fig.~\ref{introduction}(b).
To address this issue, we introduce a random walk module to refine the pseudo labels by encouraging local consistency.
Viewing each point as a node, we build a graph on the point cloud to propagate and smooth pseudo labels.  
Specifically, we apply the random walk algorithm~\cite{lovasz1993random} in the graph. 
Using distance on 3D point coordinates as a measure, we build an affinity matrix to describe the similarity between two nodes.
In the affinity matrix, closer nodes will be assigned a higher score to ensure local consistency.
Normalizing the affinity matrix, we acquire the random walk transition matrix to guide the propagation among the nodes.
Through the propagation on the graph, we obtain locally consistent pseudo scene flow labels for scene flow learning.

Our main contributions can be summarized as follows:
\begin{itemize}
	\item  
	We propose a novel self-supervised scene flow learning method in point clouds (Self-Point-Flow) to generate pseudo labels  by point matching and  perform pseudo label refinement by encouraging the local consistency of the pseudo labels;

	\item  
	Converting the pseudo label generation problem into a point matching task, we propose an optimal transport module for pseudo label generation by considering multiple clues (3D coordinates, colors and surface normals) and explicitly encouraging one-to-one matching;

	\item  
	Neighboring points in an object often share a similar movement pattern. Building a graph on the point cloud, we propose a random walk module to refine the  pseudo labels by encouraging local consistency.
	
	\item  
	Our proposed Self-Point-Flow achieves state-of-the-art performance among self-supervised learning methods.
	Our self-supervised method even performs on par with some supervised learning approaches, although we do not need any ground truth flow for training.
\end{itemize}

\section{Related Work}

\noindent\textbf{Supervised scene flow from point clouds}\quad 
Scene flow is first proposed in~\cite{vedula1999three} to represent the 3D motion of points  in a scene. 
Many works~\cite{dewan2016rigid,ushani2017learning,menze2015object,ma2019deep,quiroga2014dense,huguet2007variational,valgaerts2010joint,vogel20113d,vogel2013piecewise,vogel20153d} have been proposed to recover scene flow from multiple types of data.
Recently, directly estimating scene flow from point cloud data using deep learning has become a new research direction.
Some approaches~\cite{wu2020pointpwc,puy2020flot,gu2019hplflownet,liu2019flownet3d,behl2019pointflownet,wang2018deep} learn scene flow in point clouds in a fully supervised manner.
Puy~et~al.~\cite{puy2020flot} first introduce the optimal transport into this field.
Added into DNNs, this optimal transport module uses learned features to regress scene flow under full supervision.
Different from~\cite{puy2020flot}, we focus on unsupervised learning, and our optimal transport module leverages low-level clues to match points for pseudo label generation.

\noindent\textbf{Unsupervised scene flow from point clouds}\quad 
To circumvent the need for expensive ground truth,  some approaches~\cite{wu2020pointpwc,mittal2020just} target self-supervised learning. 
Mittal~et~al.~\cite{mittal2020just} introduce a nearest neighbor loss and an anchored cycle loss. 
Wu~et~al.~\cite{wu2020pointpwc} use the Chamfer distance~\cite{fan2017point} as the main proxy loss.
For both the nearest neighbor loss and the Chamfer distance, the nearest neighbor in the other point cloud is regarded as the corresponding point to provide supervision signals.
Unlike~\cite{mittal2020just,wu2020pointpwc}, when building correspondences, we utilize multiple descriptors as clues and  leverage global mass constraints to explicitly encourage one-to-one matching  in optimal transport.

\noindent\textbf{Unsupervised optical/scene flow from images}\quad 
Other relative topics are unsupervised optical flow from images~\cite{ren2017unsupervised,yin2018geonet,zou2018df,liu2019selflow,liu2020flow2stereo} and unsupervised scene flow from images~\cite{hur2020self,liu2019unsupervised,wang2019unos}.
In these scopes, the photometric consistency is widely used as a proxy loss to train flow estimation networks by penalizing the photometric differences.
Different from these works that directly use the differences as the supervision signal, our method produces pseudo ground truth, which enables our self-supervised method to cooperate with any point-wise loss function.

\noindent\textbf{Optimal transport}\quad 
Optimal transport has been studied in various fields, such as few-shot learning~\cite{zhang2020deepemd,zhang2020deepemdv2}, pose estimation~\cite{sarlin2020superglue}, semantic correspondence~\cite{liu2020semantic}, and etc.
Most of them embed the optimal transport into DNNs to find correspondences with learnable features.
In this paper, we apply optimal transport to self-supervised scene flow learning. 

\noindent\textbf{Random walk}\quad 
Random walk is a widely known graphical model~\cite{lovasz1993random}, which has been used in image  segmentation~\cite{bertasius2017convolutional} and  person re-ID~\cite{shen2018deep}.  Bertasius~et~al.~\cite{bertasius2017convolutional} use pixel-to-pixel relations to regularize the pixel prediction results.
Shen~et~al.~\cite{shen2018deep} use inter-image relations to improve image affinity ranking.
In this paper, based on the local consistency assumption, we focus on leveraging point-to-point relations for pseudo label smoothness and generation.

\section{Method}

\begin{figure*}[tb]
	\centering
	\includegraphics[height=4.5cm]{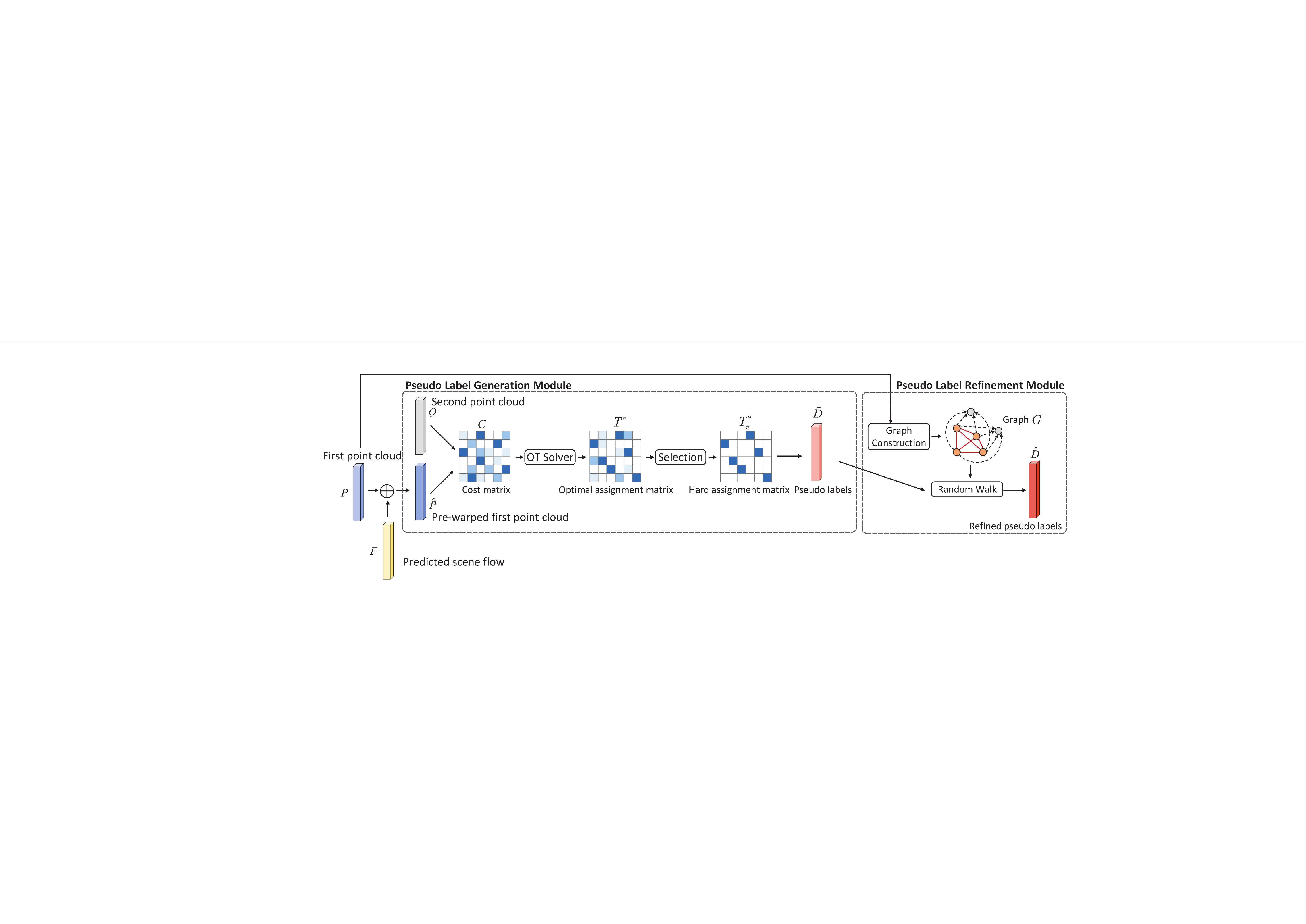}
	\caption{The pipeline of our proposed self-supervised scene flow learning method. Two parts constitute this method: a pseudo label generation module to produce initial pseudo labels by point matching and a pseudo label refinement module to improve the quality of  the pseudo labels by encouraging local consistency.}
	\label{fig_p}	
\end{figure*}

In this section, we first introduce the theory of optimal transport, and then we discuss the relationship between scene flow labels and point correspondences.
Based on the relationship, we solve the pseudo label generation problem by finding point correspondences in an optimal transport framework. 
Finally, we introduce the details of our proposed pseudo label refinement module that produces dense and locally consistent pseudo labels by the random walk theory.
The overview of our method is illustrated in Fig.~\ref{fig_p}.

\subsection{Optimal Transport Revisited}
Optimal transport problem~\cite{villani2008optimal} seeks a transport plan that moves a source distribution $\mu_s$ to a target distribution $\mu_t$ with a minimum transport cost.
In the discrete versions of this problem, $\mu_s$ and $\mu_t$ are defined as discrete empirical distributions in $ \mathbb{R}^n$.
Adapting Kantorovich’s formulation~\cite{kantorovitch1958translocation} to the discrete setting, the space of transport plans is a polytope, and the discrete optimal transport problem can be written as:
\begin{equation}
\begin{aligned}
&{ \bm U}^* =  \mathop{\arg\min}_{ \bm U \in \mathbb{R} ^{n \times n}_{+}} \sum_{ij} {\bm H}_{ij} {\bm U}_{ij} \\
&\text{s.t.} \quad {\bm U} {\bm 1}_{n} = \mu_s, \ {\bm U}^\mathsf{T} {\bm 1}_{n} = \mu_t,
\end{aligned}
\end{equation}
where $\bm H_{ij} \ge 0$ is the transport cost from sample $i$ to sample $j$, $\bm U^*$ is the optimal assignment matrix and each element $\bm U_{ij}^*$ describes the optimal amount of mass transported from sample $i$ to sample $j$.

\subsection{Pseudo Label Generation by OT }
Given two consecutive point clouds,  ${\bm P} = \{{\bm p}_i \in  \mathbb{R}^3 \}_{i=1}^{n}$ at frame $t$ and ${\bm Q} = \{{\bm q}_i \in  \mathbb{R}^3 \}_{j=1}^{n}$ at frame $t+1$, the task of point cloud scene flow estimation aims to predict the scene flow ${\bm F} = \{{\bm f}_i \in  \mathbb{R}^3 \}_{i=1}^{n}$ for point cloud $\bm P$, where each element $\bm f_i$ represents the translation of point $\bm p_i$ from frame $t$ to frame $t+1$.

Unlike fully supervised scene flow learning, where scene flow labels are available, the self-supervised scene flow learning should produce pseudo labels or design self-supervised losses for training. 
In this paper, we study how to generate effective pseudo labels for scene flow learning. \\

\noindent\textbf{Extracting pseudo labels via point matching}\quad 
Scene flow describes the motion between two consecutive point clouds.
Ideally, if no viewpoint shift and occlusions exist, following the ground truth scene flow labels~$\bm D$, the first point cloud $\bm P$ can be projected into the next frame and fully occupy the second point cloud $\bm Q$:
\begin{equation}\label{flow}
 {\bm P} + {\bm D} = {\bm {\pi}} \ {\bm Q},
\end{equation}
where ${\bm {\pi}} \in \{0, 1\}^{n \times n}$ is a permutation matrix to indicate the point correspondences between the two point clouds.
Therefore, for a pair of consecutive point clouds $\bm P$ and $\bm Q$, if we can accurately match points in the two point clouds, $i.e.$ accurately computing the permutation matrix~$\bm {\pi} $, the  correspondences derived from~$\bm {\pi} $ can help us recover the ground truth scene flow labels~$\bm D$.
In other words, we can solve the pseudo label generation problem by finding point correspondences.

When building correspondences, a straightforward  way is to directly match the points from $\bm P$ to $\bm Q$.
However, for the self-supervised scene flow estimation task, given predicted scene flow~$\bm F$, we propose a pre-warping operation to warp the first point cloud $\bm P$ by the predicted scene flow~$\bm F$, and then find correspondences by matching points from the pre-warped first point cloud, denoted as~$\bm { \hat P}$, to the second point cloud $\bm Q$.
Although the predicted scene flow is inaccurate at the beginning of the training, the predictions will be gradually improved as the training continues, which makes the matching from~$\bm { \hat P}$ to $\bm Q$ easier than the matching from $\bm {P}$ to $\bm {Q}$.
In Sec.~\ref{AS}, we show that the matching from ~$\bm { \hat P}$ to $\bm Q$ can make our self-supervised method achieve better performance.\\

\noindent\textbf{Building optimal transport problem}\quad 
Using 3D point coordinate, color, and surface normal as measures to compute the matching cost and formulating one-to-one matching as the mass equality constraints, we build an optimal transport problem from~$\bm { \hat P}$ to~$\bm Q$,
\begin{equation}\label{OT}
\begin{aligned}
&\bm T^* =  \mathop{\arg\min}_{\bm T \in \mathbb{R} ^{n \times n}_{+}} \sum_{ij} {\bm C}_{ij} \bm T_{ij} \\
&\text{s.t.} \quad \bm T {\bm 1}_{n} = {\bm \mu_{\hat p}}, \  \bm T^\mathsf{T} {\bm 1}_{n} ={\bm \mu_q}.
\end{aligned}
\end{equation}
$\bm T^*$ is the optimal assignment matrix from~$\bm { \hat P}$ to~$\bm Q$. 
$\bm C_{ij}$ is the transport cost from the $i$-th point in $\bm { \hat P}$ to the $j$-th point in~$\bm Q$. 
The transport cost $\bm C_{ij}$ is obtained by computing the pairwise difference between $\bm { \hat p_i}$ and $\bm { q_j}$ in the three measures.
The coordinate cost~$\bm C_{ij}^d$ and the color cost~$\bm C_{ij}^c$ are defined on a Gaussian function: 
\begin{small}
\begin{equation}
\setlength{\abovedisplayskip}{3pt}
\setlength{\belowdisplayskip}{3pt}
{\bm C}_{ij}^{d} = 1- {\rm \exp}  (- \frac{\| \bm{\hat p}_i - \bm{q}_j \|^2}{2\theta_d^2}),
\end{equation}
\end{small}
\begin{small}
\begin{equation}
\setlength{\abovedisplayskip}{3pt}
\setlength{\belowdisplayskip}{3pt}
{\bm C}_{ij}^{c} = 1- {\rm \exp}  (- \frac{\| \bm{k}_{{\hat p},i}^{c} - \bm{k}_{q,j}^{c} \|^2}{2\theta_c^2}),
\end{equation}
\end{small}
where $\|\cdot\|$ denotes the $L^2$ norm of a vector, $\theta_d$ and $\theta_c$ are user defined parameters, $\bm {\hat p}_i$ and $\bm {q}_j$ represent the coordinates of the $i$-th point and the $j$-th point, $\bm{k}_{{\hat p},i}^{c}$ and $\bm{k}_{q,j}^{c}$ are the colors of the two points.
The surface normal cost is calculated using the cosine similarity:
\begin{small}
\begin{equation}
{\bm C}_{ij}^{s} = 1- \frac{ \| (\bm{k}_{{\hat p}, i}^{s})^\mathsf{T}  \bm{k}_{{q}, j}^{s} \| }{\| \bm{k}_{{\hat p}, i}^{s} \| \cdot \| \bm{k}_{{q}, j}^{s} \|},
\end{equation}
\end{small}
where $\bm{k}_{{\hat p},i}^{s}$ and $\bm{k}_{q,j}^{s}$ are the surface normals of the two points.
The final transport cost is the sum of the three individual costs:
\begin{small}
	\begin{equation}
	\setlength{\abovedisplayskip}{3pt}
	\setlength{\belowdisplayskip}{3pt}
	{\bm C}_{ij} = {\bm C}_{ij}^{d} + {\bm C}_{ij}^{c}  + {\bm C}_{ij}^{s}. 
	\end{equation}
\end{small}

In order to encourage one-to-one matching, in the equality constraints of Eq.~\ref{OT}, we set ${\bm \mu_{\hat p}} = \frac{1}{n}{\bm 1}_n$~and~${\bm \mu_q} = \frac{1}{n}{\bm 1}_n$. 
In this case, the row sum and the column sum of assignment matrix $\bm T$ are constrained to be a uniform distribution, which will alleviate the many-to-one matching problem.\\

\noindent\textbf{Efficiently solving with the Sinkhorn algorithm}\quad 
To efficiently solve the optimal transport problem, we smooth the above problem with an entropic regularization term:
\begin{equation}\label{smooth_OT}
\begin{aligned}
&\bm T^* =  \mathop{\arg\min}_{\bm T \in \mathbb{R} ^{n \times n}_{+}} \sum_{ij} {\bm C}_{ij}  \bm T_{ij}
+ \varepsilon  \bm T_{ij} (\log{ \bm T_{ij}} - 1)\\
&\text{s.t.} \quad  \bm T {\bm 1}_{n} = {\bm \mu_{\hat p}}, \  \bm T^\mathsf{T} {\bm 1}_{n} ={\bm \mu_q}.
\end{aligned}
\end{equation}
$\varepsilon$ is the regularization parameter. The Sinkhorn algorithm~\cite{cuturi2013sinkhorn} can be employed to solve this entropy-regularized formulation. 
The details are presented in Algorithm~\ref{alg1}. \\

\begin{algorithm}[tb]
	\caption{Optimal transport }
	\label{alg1}
	\hspace*{0.02in}{\bf Input:} 
	Transport cost matrix $\bm C$; hyperparameter $\varepsilon$, iteration number $ L_{o}$;\\
	\hspace*{0.02in}{\bf Output:} 
	Optimal transport matrix ${T}^*$ ;\\
	\hspace*{0.02in}{\bf Procedure:} 
	\begin{algorithmic}[1]
		\STATE {${\bm K}   \gets  {\rm \exp}(-{\bm C} / \varepsilon)$;}
		\STATE {${\bm \mu_{\hat p}}   \gets \frac{1}{n}{\bm 1}_n$, ${\bm \mu_{q}}   \gets \frac{1}{n}{\bm 1}_n$, ${\bm a}  \gets \frac{1}{n}{\bm 1}_n$;}\\
		\FOR{ $l = 1,...,L_{o}$} 
		\STATE {${\bm b}  \gets {\bm \mu_{q}}  /  ({\bm K}^\mathsf{T} {\bm a})$;}
		\STATE {${\bm a}  \gets {\bm \mu_{\hat p}}  /  ({\bm K} {\bm b})$;}
		\ENDFOR
		\STATE{${T}^* \gets  \operatorname{diag}(\bm a) {\bm K}  \operatorname{diag}(\bm b)  $.} 	
	\end{algorithmic}
\end{algorithm}

\noindent\textbf{Selecting hard correspondences and generating pseudo labels from assignment matrix}\quad 
The optimal transport plan $\bm T^*$ derived from Algorithm~\ref{alg1} is a soft assignment matrix, where~$\bm T^*_{ij} \in [0, 1]$.
To obtain hard correspondences from $\bm{\hat P}$ to  $\bm{Q}$, in each row of  $\bm T^*$, we set the element with the maximum value to 1 and the remaining elements to 0 so that the point with the highest transport score is selected as the unique corresponding point in this row.
The produced hard assignment matrix is donated as $\bm T^*_{\pi}$.
According to the Eq.~\ref{flow}, we have the pseudo scene flow labels~$\bm {\tilde {D}}$:
\begin{equation}
{\bm {\tilde {D}}} = {\bm T^*_{\pi}}  {\bm Q}  -   {\bm P}.
\end{equation}
Removing some invalid pseudo scene flow labels with too large displacements (larger than 3.5$m$), we obtain a set of valid pseudo labels~$\bm {\tilde{D}}_M$ for the valid labeled points~$\bm P_M$.
And the remaining points without valid pseudo labels are denoted as~$\bm P_S$.

\subsection{Pseudo Label Refinement by Random Walk}

The optimal transport module generates pseudo labels by point-wise matching but lacks in capturing the local relations among neighboring points, resulting in conflicting pseudo labels in each local region.
Moreover, after the pseudo label generation module, there are still some points without valid pseudo labels.
To address the issues, we propose a pseudo label refinement module to encourage the local consistency of pseudo labels and infer new pseudo labels for those unlabeled points.\\

\begin{figure}[tb]
	\centering
	\includegraphics[height=3.0cm]{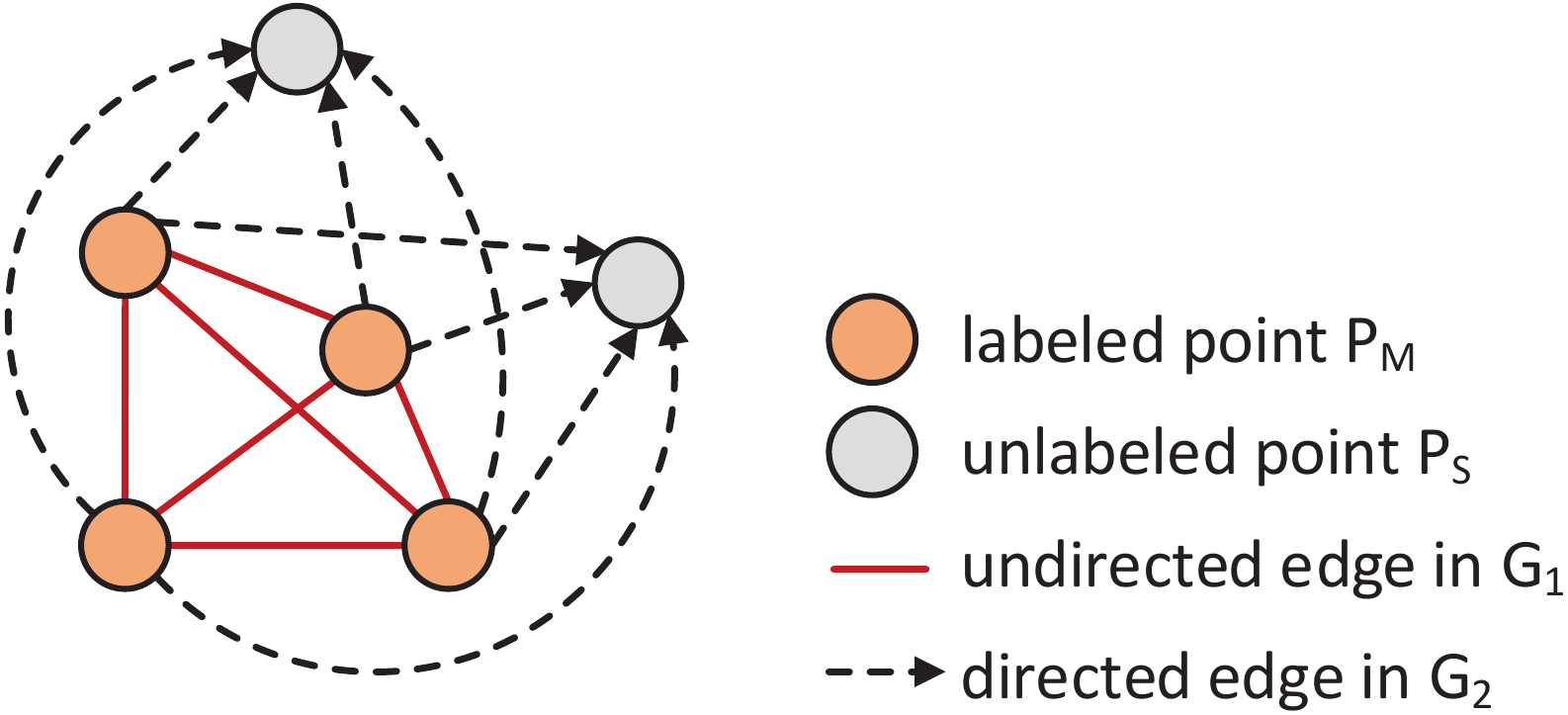}
	\caption{Illustration of the graph~$G$. This graph consists of two subgraphs: a fully-connected undirected subgraph for pseudo label smoothing on labeled points and a directed subgraph for new pseudo label generation on unlabeled points.}
	\label{fig_g}	
\end{figure}

\noindent\textbf{Building graph on the point cloud}\quad 
Viewing each point as a node, we build a graph $G(V,E)$ on the first point cloud~$\bm P$, shown in Fig.~\ref{fig_g}.
According to the labeling state of each point, the nodes are separated into two sets, the labeled nodes associated with~$\bm P_M$  and the unlabeled nodes associated with~$\bm P_S$.
$V_{m} = \{1, 2, ..., n_m \}$ and $V_{s} = \{1, 2, ..., n_s \}$ represent the labeled node set and the unlabeled node set, respectively.
$n_m$ and $n_s$ are the number of nodes in the two sets. 
Subsequently, the entire graph $G(V,E)$ can be divided into two suggraphs: 
1) a fully-connected undirected subgraph~$G_1(V_m,E_m)$ on labeled nodes~$V_m$ to smooth the pseudo labels of the labeled points;
2) a directed subgraph~$G_2(V,E_s)$  from labeled nodes~$V_m$ to unlabeled nodes~$V_s$ to generate new pseudo labels for the unlabeled points.  
In this procedure,  the pseudo labels of the unlabeled points are entirely dependent on those of the labeled points.
Therefore, we can first propagate pseudo labels on the undirected subgraph and then on the directed subgraph.
The propagation operation can be achieved by the random walk algorithm~\cite{lovasz1993random}.\\

\noindent\textbf{Propagating on the undirected subgraph}\quad 
The fully-connected undirected subgraph is constructed to improve the local consistency of pseudo labels for the labeled point set~$\bm P_{M} $.
The random walk operation on this subgraph can be modeled with a~$n_m \times n_m$ transition matrix~$\bm A^{1}$.
$\bm A^{1}_{ij}~\in~[0, 1]$ denotes the transition probability between $i$-th and $j$-th nodes with constraints~$\sum_j \bm A^{1}_{ij} = 1$ for all~$j$.

To encourage the local consistency, we use the nearness among nodes as the measure to build the transition matrix so that the closer nodes will be assigned a higher transition probability.
Firstly, we denote a symmetric~$n_m \times n_m$ affinity matrix~$ \bm W^{1}$, where each element~$\bm W^{1}_{ij}$ describes how near the nodes $i$ and $j$ are,
\begin{small}
	\begin{equation}\label{W}
		\setlength{\abovedisplayskip}{3pt}
	\setlength{\belowdisplayskip}{3pt}
	{\bm W}_{ij}^{1} = {\rm \exp}  (- \frac{\| \bm p_i - \bm p_j \|^2}{2\theta_r^2}),
	\end{equation}
\end{small}
where $\theta_r$ is a hyperparameter, $\bm p_i$ and $\bm p_j$ are point coordinates associated with nodes $i$ and $j$.
Then, we normalize the affinity matrix~$ \bm W^{1}$ to obtain the transition matrix~$\bm A^{1}$, where each element~$ \bm A_{ij}^{1}$ is written as:
\begin{small}
	\begin{equation}\label{A}
	{\bm A}_{ij}^{1} = \frac{{\bm W}_{ij}^{1}}{\sum_{j \neq i}{\bm W}_{ij}^{1}}.
	\end{equation}
\end{small}
The $t$-th iteration of random walk refinements on the pseudo labels can be expressed as
\begin{equation}
{\bm {\tilde{D} }_M^{(t)}}   =  \alpha {\bm A^{1}} {\bm {\tilde{D}}_M^{(t-1)}} + (1 - \alpha)  {\bm {\tilde{D}}_M^{(0)}},
\end{equation}
where ${\bm {\tilde {D}}}_M^{(0)}$ are the initial pseudo labels derived from the pseudo label generation module, 
${\bm {\tilde {D}}}_M^{(t-1)}$ are the refined labels after $t-1$ random walk steps,
and $\alpha$ is a parameter $[0, 1]$ to control the tradeoff between the random walk refinements and the initial values. 

When applying the random walk procedure until convergence, $i.e.$  $t = \infty$, according to~\cite{bertasius2017convolutional,shen2018deep}, the final random walk refinements can be written as
\begin{equation}
{\bm {\tilde{D} }_M^{(\infty)}}   =  (1 - \alpha) ({\bm I} -\alpha {\bm A^{1}})^{-1}  {\bm {\tilde{D}}_M^{(0)}},
\end{equation}
where ${\bm I}$ is the identity matrix.
After $L_r$ random walk steps, we treat the produced random walk refinements as the refined pseudo labels of the labeled points, ${\bm {\widehat D}}_M = \bm {\tilde {D}}_M^{(L_r)}$.\\

\noindent\textbf{Propagating on the directed subgraph}\quad 
The undirected subgraph is built to infer new pseudo labels for the unlabeled point set~$\bm P_S$ based on the refined pseudo labels of the labeled point set,~${\bm {\widehat D}}_M$.
Similar to the propagation process on the undirected subgraph, we first define a $n_s \times n_m$ affinity matrix $\bm W^{2}$ to describe the nearness between each point in~$\bm P_S$ and each point in~$\bm P_S$.
Then we obtain a $n_s \times n_m$ transition matrix~$ \bm A^{2}$ by normalizing the  affinity matrix~$\bm W^{2}$. 
The calculation of~$\bm W^2$ and~$\bm A^2$ is the same as that of~$\bm W^1$ and~$\bm A^1$, shown in Eq.~\ref{W} and Eq.~\ref{A}. 
Based on the transition matrix~$ \bm A^{2}$ and the refined pseudo labels ${\bm {\widehat D}}_M$, we obtain the new pseudo labels~${\bm {\widehat D}}_S$ for the unlabeled points:
\begin{equation}
{\bm {\widehat D}}_S =\bm A^{2}  {\bm {\widehat D}}_M.
\end{equation}

\noindent\textbf{Training with pseudo labels}\quad 
Combining the refined pseudo labels ${\bm {\widehat D}}_M$ and the new pseudo labels~${\bm {\widehat D}}_S$, we obtain the final refined pseudo labels~$\bm {\widehat D}$ for the entire point cloud~$\bm P$ in self-supervised learning.
The training loss in our self-supervised learning method can be computed by:
\begin{equation}
Loss =f_{loss}({\bm {\widehat D}}, {\bm {F}}), 
\end{equation}
where $f_{loss}$ is any per-point loss function, $\bm F$ is the predicted scene flow.
Specifically, we set  $f_{loss}$ to per-point $L_2$-norm loss function for scene flow learning in this paper.

\section{Experiments}
We first compare our method with two state-of-the-art self-supervised scene flow estimation methods in Sec.~\ref{CS}.
Then, we compare our self-supervised models with state-of-the-art fully-supervised models in Sec.~\ref{CF}. 
Finally, we conduct ablation studies to analyze the effectiveness of each component in  Sec.~\ref{AS}.
In this section, we adopt the FlowNet3D~\cite{liu2019flownet3d} as our default scene flow estimation model with only point coordinates as input. Experiments will be  conducted on FlyingThings3D~\cite{mayer2016large} and KITTI 2015~\cite{menze2015object,menze2015joint} datasets.
Point clouds are not directly provided in the two original datasets.
Following~\cite{puy2020flot}, we denote the two processed point cloud datasets provided by~\cite{gu2019hplflownet} as FT3D$_{\rm s}$ and KITTI$_{\rm s}$.
And we denote the two processed datasets provided by~\cite{liu2019flownet3d} as FT3D$_{\rm o}$ and KITTI$_{\rm o}$.

\noindent\textbf{Evaluation metrics.}\quad
We adopt four evaluation metrics used in \cite{liu2019flownet3d}, \cite{gu2019hplflownet}, \cite{puy2020flot}.
Let $\bm Y$ denote the predicted scene flow, and $\bm D$ be the ground truth scene flow. The evaluate metrics are computed as follows.
\textbf{EPE}(m): the main metric, $\| \bm Y^* - \bm Y_{gt}\|_2$ average over each point.
\textbf{AS}(\%): the percentage of points whose EPE $<$ 0.05m or relative error $< 5\%$.
\textbf{AR}(\%): the percentage of points whose EPE $<$ 0.1m or relative error $< 10\%$.
\textbf{Out}(\%): the percentage of points whose EPE $>$ 0.3m or relative error $> 10\%$.

\subsection{Comparison with self-supervised methods}\label{CS}
\noindent\textbf{Comparison with PointPWC-Net~\cite{wu2020pointpwc}.}\quad
Wu et al.~\cite{wu2020pointpwc} introduce Chamfer distance, smoothness constraint, and Laplacian regularization for self-supervised learning.
Following the experimental settings of~\cite{wu2020pointpwc}, we first train the FlowNet3D model with our self-supervised method on FT3D$_{\rm s}$ and then evaluate on FT3D$_{\rm s}$ and KITTI$_{\rm s}$.
During training, we use the whole training set in FT3D$_{\rm s}$ for training.
Besides, we also try to add the cycle-consistency regularization~\cite{liu2019flownet3d} into our training loss.
The detailed experimental setting could be found in supplementary.

The results are shown in Table~\ref{table_1}. Our method outperforms self-supervised PointPWC-Net~\cite{wu2020pointpwc} on all metrics and shows significantly better generalization ability on KITTI, although the network capacity of our used FlowNet3D is  worse than that of their PointPWC-Net.
Adding the cycle-consistency regularization to the loss function, our model gains a further improvement.\\

\begin{table}\footnotesize
	
	\caption{ Evaluation results on FlyingThings3D and KITTI datasets using the process point cloud data provided by~\cite{gu2019hplflownet}.
		\textit{Full} means fully-supervised training, \textit{Self}  means self-supervised training. 
	$\dagger$ means that we add a cycle-consistency regularization~\cite{liu2019flownet3d} into the training loss.
    Without using ground truth flow, our self-supervised method outperforms PointPWC-Net on the two datasets and even performs on par with some supervised approaches.}
	\label{table_1}
	\renewcommand\arraystretch{1.05}	
	\centering	
	
	\begin{tabular}{l@{\hskip 0.1cm}|l@{\hskip 0.1cm}|c@{\hskip 0.1cm}|c@{\hskip 0.1cm}c@{\hskip 0.1cm}c@{\hskip 0.1cm}c}
	 	\Xhline{1.2pt}
		{Dataset} & {Method} & {Sup.} & {EPE$\downarrow$} & {AS$\uparrow$} & {AR$\uparrow$} & { Out$\downarrow$}\\ 
		\hline\multirow{8}{0.0cm}{FT3D$_{\rm s}$} 
		&PointPWC-Net~\cite{wu2020pointpwc} & {\textit{Self}}& 0.1213 & 32.39  &  67.42 & 68.78  \\
		&Ours & {\textit{Self}}&0.1208	& 36.68	& 70.22	& 65.35 \\
		&Ours$^\dagger$ & {\textit{Self}}& \bf 0.1009& \bf 42.31	& \bf 77.47	& \bf 60.58  \\
		\cline{2-7}
		&SPLATFlowNet~\cite{su2018splatnet} &{\textit{Full}}&   0.1205 & 41.97 &71.80 &61.87  \\
		&original BCL~\cite{gu2019hplflownet} &{\textit{Full}}&   0.1111 & 42.79 &  75.51 & 60.54    \\
		&FlowNet3D~\cite{liu2019flownet3d} & {\textit{Full}}&  0.0864&   47.89&  83.99 &  54.64 \\
		&HPLFlowNet~\cite{gu2019hplflownet} &{\textit{Full}}&   0.0804 & 61.44  & 85.55 & 42.87   \\
		&PointPWC-Net~\cite{wu2020pointpwc} & {\textit{Full}}&  0.0588 & 73.79 & 92.76 &34.24  \\
		\hline\multirow{8}{0.0cm}{KITTI$_{\rm s}$} 
		&PointPWC-Net~\cite{wu2020pointpwc} & {\textit{Self}}&  0.2549 & 23.79 & 49.57 & 68.63   \\
		&Ours & {\textit{Self}}&  0.1271	& 45.83	&77.77	&41.44   \\
		&Ours$^\dagger$  & {\textit{Self}}&  \bf 0.1120	& \bf 52.76	& \bf 79.36	&\bf 40.86  \\
		\cline{2-7}
		&SPLATFlowNet~\cite{su2018splatnet} &{\textit{Full}}&    0.1988 & 21.74 & 53.91 &65.75   \\
		&original BCL~\cite{gu2019hplflownet} &{\textit{Full}}&   0.1729 & 25.16 & 60.11 & 62.15 \\
		&FlowNet3D~\cite{liu2019flownet3d} & {\textit{Full}}&  0.1064& 50.65  & 80.11  &  40.03 \\
		&HPLFlowNet~\cite{gu2019hplflownet} &{\textit{Full}}&   0.1169& 47.83 & 77.76& 41.03   \\
		&PointPWC-Net~\cite{wu2020pointpwc} & {\textit{Full}}&  0.0694 & 72.81 & 88.84 &26.48   \\
		\Xhline{1.2pt}
	\end{tabular}
\end{table}

\noindent\textbf{Comparison with JGF~\cite{mittal2020just}.}\quad 
Mittal et al.~\cite{mittal2020just} propose a nearest neighbor loss and an anchored cycle loss for self-supervised training.
In~\cite{mittal2020just},  they split the KITTI$_{\rm o}$ into two sets, 100 pairs of point clouds for training, denoted as KITTI$_{\rm v}$, and  the remaining 50 pairs for testing, denoted as KITTI$_{\rm t}$.
Moreover, they also leverage an additional real-world outdoor point cloud dataset, nuScenes~\cite{caesar2020nuscenes}, to augment their training data.
In~\cite{mittal2020just}, all networks are initialized with a  Flownet3D model~\cite{liu2019flownet3d} pre-trained on FlyingThing3D.

In our experiment, we use the raw data from KITTI to produce point clouds as our training data.
Because the point clouds in KITTI$_{\rm o}$ belong to 29 scenes in KITTI, to avoid the overlap of training data and test data, we produce training point clouds from the remaining 33 scenes.  
Extracting a pair of point clouds at every five frames, we build a self-supervised training set containing 6,068 pairs, denoted as KITTI$_{\rm r}$.
For comparison, we test our model on the same KITTI$_{\rm t}$ with 50 test pairs.
In each test pair, our model is evaluated by processing 2,048 random points. 
The detailed experimental setting is in supplementary.

\begin{figure}[tb]
	\vspace{-0.3cm}
	\centering
	\includegraphics[height=3.8cm]{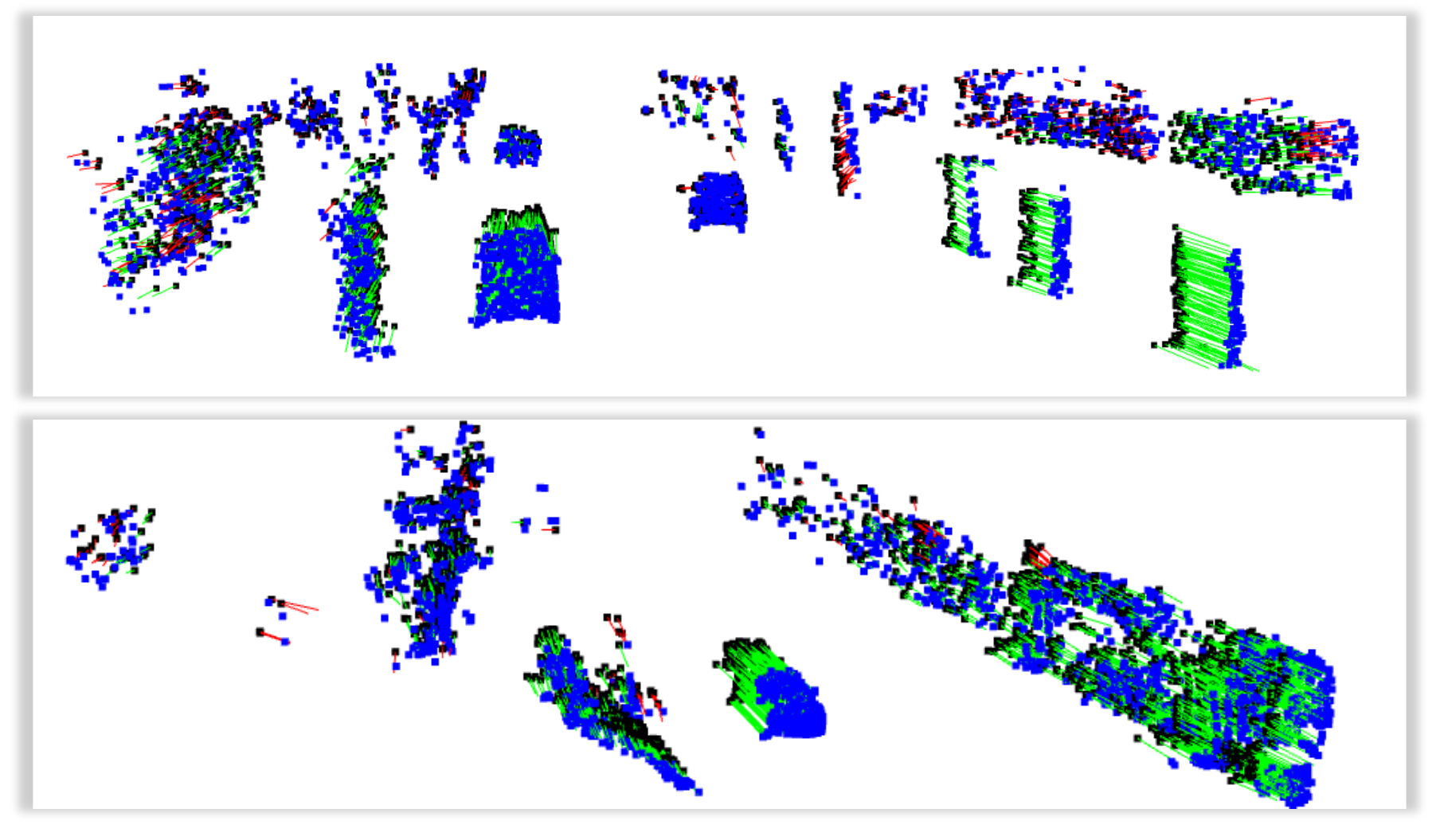}
	\caption{Pseudo ground truth of some examples. Blue points are the first point cloud.  Black points are the second point cloud.
				Green line represents the correct pseudo ground truth measured by \textbf{AR}.
				Red line represents the wrong pseudo ground truth.
	}
	\label{fig_sample}	
\end{figure}

The results are shown in Table~\ref{table_2}.
Our model trained on KITTI$_{\rm r}$ outperforms their model by 18.3\% in \textbf{EPE}, which is pre-trained on FT3D and then trained on KITTI$_{\rm v}$, although training from scratch is much more challenging than fine-tuning a pre-trained model  for self-supervised learning.
After further fine-tuned on KITTI$_{\rm v}$, our model achieves comparable performance to their model, which is pre-trained on FT3D and then trained on nuScenes and KITTI$_{\rm v}$.
When using the parameters of self-supervised models as initial weights and performing fully-supervised training on  KITTI$_{\rm v}$, our model outperforms theirs on all metrics.
Fig.~\ref{fig_sample} displays our produced pseudo ground truth for some examples in KITTI$_{\rm v}$.

\begin{table}\scriptsize
	
	\caption{ Evaluation results on KITTI$_{\rm t}$ according to the test settings of Mittal et al.~\cite{mittal2020just}.
	$\ddagger$ means a fully-supervised fine-tuning on KITTI$_{\rm v}$.}
	\label{table_2}
	\renewcommand\arraystretch{1.2}	
	\centering	
	
	\begin{tabular}{c@{\hskip 0.1cm}|c@{\hskip 0.1cm}|l@{\hskip 0.1cm}|c@{\hskip 0.1cm}c@{\hskip 0.1cm}c}
		\Xhline{1.2pt}
		{Method} &{Pre-trained} &{Training data} & {EPE$\downarrow$} & {AS$\uparrow$} & {AR$\uparrow$} \\ \hline
		\cite{mittal2020just} &$\checkmark$ & FT3D$_{\rm o}$  + KITTI$_{\rm v}$ &0.1260 & 32.00 & 73.64  \\
		\cite{mittal2020just} &$\checkmark$ & FT3D$_{\rm o}$ + nuScenes + KITTI$_{\rm v}$ &0.1053 & 46.48 & 79.42 \\
		Ours & &KITTI$_{\rm r}$ &  0.1029&	35.68	& 68.55  \\
		Ours & &KITTI$_{\rm r}$ + KITTI$_{\rm v}$ &  0.0895	& 41.74	& 75.01   \\

		\cite{mittal2020just} &$\checkmark$ & FT3D$_{\rm o}$ + nuScenes + KITTI$_{\rm v}$$^\ddagger$   &0.0912 & 47.92 & 79.63 \\
		Ours & &KITTI$_{\rm r}$ + KITTI$_{\rm v}$$^\ddagger$ & \bf 0.0720	& \bf 50.12	 & \bf 82.38   \\
		\Xhline{1.2pt}
	\end{tabular}
\end{table}

\begin{figure*}[tb]
	\centering
	\includegraphics[height=5.0cm]{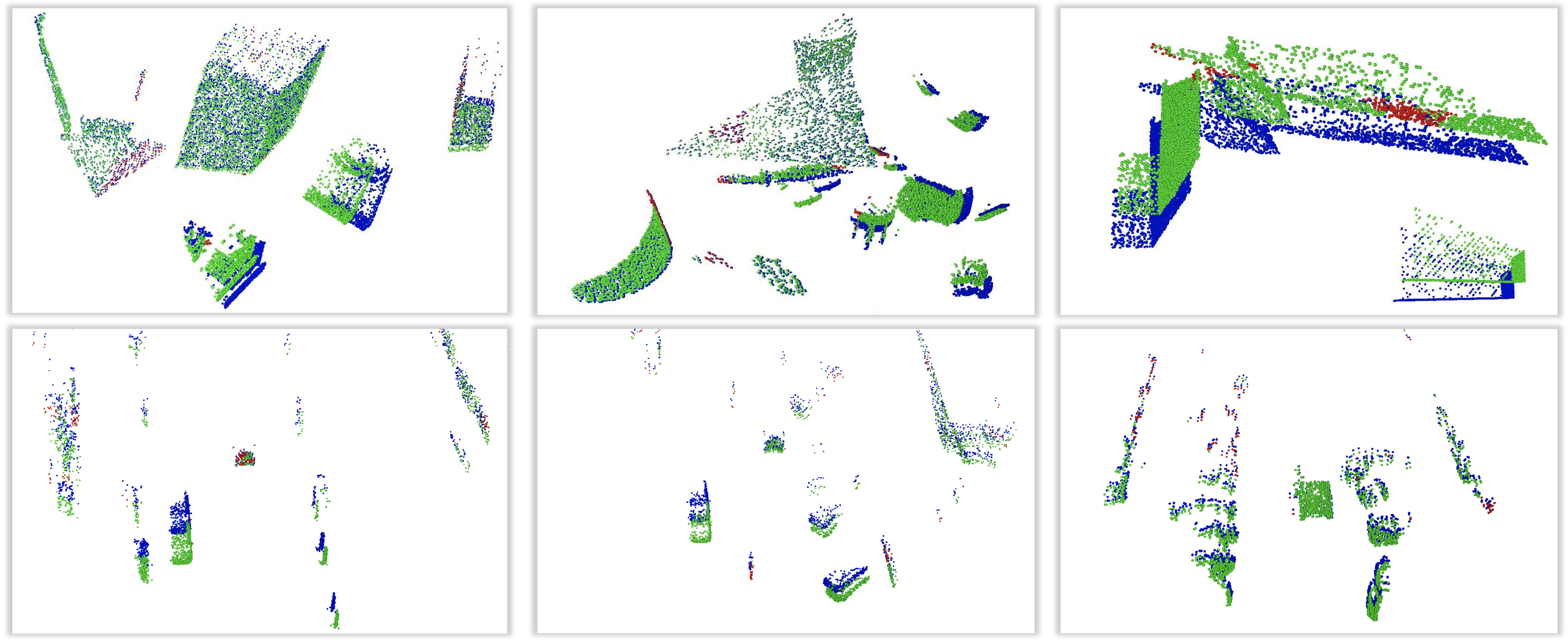}
	\caption{Qualitative results on FlyingThings3D (top) and KITTI (bottom).   Blue points are the first point cloud $\bm P$.   Green points are the points warped by the correctly predicted scene flow. The predicted scene flow belonging to \textbf{AR} is regarded as a correct prediction.
	For the points with incorrect predictions, we use the ground truth scene flow to warp them and the warped results are shown as red points.}
	\label{fig_example}	
\end{figure*}

\subsection{Comparison with fully-supervised methods}\label{CF}

In Table~\ref{table_1}, we compare our self-supervised model  with some fully-supervised methods, which are also trained on FT3D$_{\rm s}$ and tested on FT3D$_{\rm s}$ and KITTI$_{\rm s}$.
As shown in Table~\ref{table_1}, adding a cycle-consistency regularization, our self-supervised method outperforms SPLATFlowNet~\cite{su2018splatnet} on FT3D$_{\rm s}$ and generalizes better  on KITTI$_{\rm s}$ than  SPLATFlowNet~\cite{su2018splatnet}, original BCL~\cite{gu2019hplflownet}, and HPLFlowNet~\cite{gu2019hplflownet}, although we do not use any ground truth flow for training.
Qualitative results are shown in Fig.~\ref{fig_example}.

In Table~\ref{table_3}, using KITTI$_{\rm o}$ as test set, we compare our self-supervised model trained on KITTI$_{\rm r}$ with some fully-supervised methods trained on  FT3D$_{\rm o}$, following the test procedure of FLOT~\cite{puy2020flot}.
Despite using the same scene flow estimation model, our self-supervised FlowNet3D trained on KITTI$_{\rm r}$ outperforms supervised FlowNet3D~\cite{liu2019flownet3d} trained on FT3D$_{\rm o}$ by  39.3\% in the metric of \textbf{EPE}.
It demonstrates that, for the FlowNet3D model, self-supervised learning on KITTI with our method is much more effective than supervised learning on FlyingThings3D in real-world scenes.
Furthermore, as shown in Table~\ref{table_3}, our self-supervised method has achieved a close performance to the state-of-the-art supervised method, FLOT~\cite{puy2020flot}, on KITTI$_{\rm o}$ dataset.
Fig.~\ref{fig_example} provides some example results.
	
\begin{table}\footnotesize
	
	\caption{Evaluation results on KITTI$_{\rm o}$. Without using ground truth flow, our self-supervised method outperforms supervised FlowNet3D~\cite{liu2019flownet3d} and achieves comparable performance to the state-of-the-art supervised method, FLOT~\cite{puy2020flot}.}
	\label{table_3}
	\renewcommand\arraystretch{1.2}	
	\centering	
	
	\begin{tabular}{l@{\hskip 0.1cm}|c@{\hskip 0.2cm}|l@{\hskip 0.2cm}|c@{\hskip 0.2cm}c@{\hskip 0.2cm}c@{\hskip 0.2cm}c}
		\Xhline{1.2pt}
		{Method} & {Sup.}& {Training data}&{EPE$\downarrow$} & {AS$\uparrow$} & {AR$\uparrow$} & {Out$\downarrow$} \\ \hline
		FlowNet3D~\cite{liu2019flownet3d}&{\textit{Full}} & FT3D$_{\rm o}$   &0.173  & 27.6 & 60.9  & 64.9 \\
		FLOT~\cite{puy2020flot}&{\textit{Full}} &  FT3D$_{\rm o}$  & 0.107 & \bf 45.1 & \bf 74.0 & \bf 46.3\\
		Ours&{\textit{Self}} &  KITTI$_{\rm r}$  & \bf 0.105 &	41.7	& 72.5 & 50.1 \\
		\Xhline{1.2pt}
	\end{tabular}
\end{table}	
	
\subsection{Ablation studies}\label{AS}
In this section, we conduct ablation studies to analyze the effectiveness of each component.
All models are trained on KITTI$_{\rm r}$ and evaluated on KITTI$_{\rm o}$.

\noindent\textbf{Ablation study for  pseudo label generation module.}\quad 
In this module, for good point matching, we adopt color and surface normal as additional measures to build the transport cost matrix and establish the global constraints to enforce one-to-one marching.
To verify the effectiveness of our module, we design a baseline method, named greedy search, which directly finds the point with the lowest transport cost in another point cloud as the corresponding point without any constraints.

Firstly, we analyze the impact of the color measure and the surface normal measure.
As shown in Table~\ref{table_A1}, for both greedy search method and optimal transport method, adding color and surface normal as measures can boost \textbf{AS} by around 10 to 17 points. 
Compared with the original optimal transport with only 3D point coordinate as a measure, our proposed pseudo label generation module brings a 139\% improvement on \textbf{AS}, which demonstrates the discriminative ability of color and surface normal in finding correspondences.

Secondly, we analyze the impact of the global constraints.
As shown in Table~\ref{table_A1}, for all three kinds of measure combinations, adding the global constraints can increase \textbf{AS} by about 5 to 9 points, which means that addressing the many-to-one problem in point matching can greatly improve the quality of produced pseudo labels.

Thirdly, we compare different matching strategies in our module.
In our optimal transport, we search for correspondences by matching from the pre-warped first point cloud to the second point cloud, denoted as $\bm { \widehat P}~\to~\bm Q$, and regard the point with the highest transport score as the corresponding point, denoted as \textbf{Hard matching}.
To evaluate the effectiveness of our matching strategy, as shown in Table~\ref{table_A2}, we design two baseline methods: 1) baseline1 matches from the first point cloud to the second one, denoted as $\bm { P}~\to~\bm Q$, 
2) baseline 2 produces a soft corresponding point by using the transport score as the weight to perform a weighted summation of all candidate points. 
This process is denoted as \textbf{Soft matching}.
As shown in Table~\ref{table_A2}, our method outperforms baseline1 and baseline2 by about 10 points on \textbf{AS}, which demonstrates the effectiveness of our matching strategy.\\

\begin{table}\scriptsize
	\caption{Ablation study for color measure, surface normal measure  and the matching constraints in our pseudo label generation module (\textbf{PLGM}). }
	\label{table_A1}
	\renewcommand\arraystretch{1.2}	
	\centering	
	
	\begin{tabular}{l@{\hskip 0.1cm}|c@{\hskip 0.1cm}c@{\hskip 0.1cm}c@{\hskip 0.1cm}c@{\hskip 0.0cm}|r}
		\Xhline{1.2pt}
		{Method}&{Coordinate} & {Color} & {Norm} & {Constraint} 
		& { AS$\uparrow$} \\\hline 
		Greedy search (Baseline) &$\checkmark$ & & &  & 1.85 \\
		+ Color&$\checkmark$ & $\checkmark$& & & 11.25\\
		+ Color + Norm&$\checkmark$ &$\checkmark$ &$\checkmark$ & & 18.89 \\
		\hline
		Optimal Transport&$\checkmark$ & & &$\checkmark$   & 10.19 \\
		+ Color &$\checkmark$ &$\checkmark$ & & $\checkmark$ & 21.51 \\
		+ (Color + Norm)/Our \textbf{PLGM}&$\checkmark$ &$\checkmark$ &$\checkmark$ & $\checkmark$ & \bf 24.36 \\
		\Xhline{1.2pt}
	\end{tabular}
\end{table}

\begin{table}\scriptsize

	\caption{Ablation study for  different matching strategies in our pseudo label generation module (\textbf{PLGM}).
		$\bm P~\to~\bm Q$:~match from the first point cloud to the second point cloud. 
		$\bm { \widehat P}~\to~\bm Q$:~match from the pre-warped first point cloud to the second point cloud.
		\textbf{Soft matching}:~produce labels by soft correspondences.
		\textbf{Hard matching}:~produce labels by hard correspondences. }
	\label{table_A2}
	\renewcommand\arraystretch{1.2}	
	\centering	
	
	\begin{tabular}{l@{\hskip 0.10cm}|c@{\hskip 0.1cm}c@{\hskip 0.1cm}c@{\hskip 0.1cm}c@{\hskip 0.0cm}|r}
		\Xhline{1.2pt}
		{Method}&{$\bm P \to \bm Q$} & {$\bm { \widehat P}\to \bm Q$} & {\textbf{Soft matching}} & {\textbf{Hard matching}} 
		& { AS$\uparrow$} \\\hline 
		Baseline1&$\checkmark$ & & & $\checkmark$ & 12.52 \\
		Baseline2& &$\checkmark$ &$\checkmark$ & & 13.16 \\
		Our \textbf{PLGM}& & $\checkmark$ & &$\checkmark$   & \bf 24.36 \\
		\Xhline{1.2pt}
	\end{tabular}
\end{table}

\noindent\textbf{Ablation study for  pseudo label refinement module.}\quad 
This module employs random walk operations to improve the local consistency of pseudo labels.
In this module, we build two subgraphes: an undirected one for label smoothness and a directed one for new label generation in unlabeled points.
To verify the effectiveness of our module, we design a naive smoothing unit (\textbf{NS}) that finds neighboring points by KNN search and outputs the average label of the neighboring points as the refined label.   
As show in Table~\ref{table_A3}, smoothing labels by random walk operation on the undirected subgraph (\textbf{UG}) improves \textbf{AS} from 24.36 to 40.88.
And the improvement from \textbf{UG} is significantly greater than that from the naive smoothing unit (\textbf{NS}).
By further generating new labels for the unlabeled points via random walk operation on the directed subgraph (\textbf{DG}), we achieve another improvement on \textbf{AS} by 0.86.
The great improvement demonstrates the effectiveness of our pseudo label refinement module.
And the impact of different random walk steps on our method is shown in Table~\ref{table_A4}.

\noindent\textbf{Time consumption of our pseudo-label generation process.}\quad 
To process a scene containing 2,048 points in KITTI$_{\rm r}$, the pseudo label generation module takes about 3.2ms and the pseudo label refinement module takes about 75.6ms on a single 2080ti GPU.
Thus, the total time consumption for a scene is 78.8ms.

\begin{table}\footnotesize
	\caption{Ablation study for  our pseudo label refinement module (\textbf{PLRM}). 
	\textbf{NS}:~naive smoothing unit. 
	\textbf{UG}:~smooth labels by random walk operation on the undirected subgraph.
    \textbf{DG}:~generate new labels by random walk operation on  the directed subgraph. }
	\label{table_A3}
	\renewcommand\arraystretch{1.0}	
	\centering	
	
	\begin{tabular}{l@{\hskip 0.30cm}|c@{\hskip 0.5cm}c@{\hskip 0.5cm}c@{\hskip 0.5cm}|r}
		\Xhline{1.2pt}
		{Method}&{\textbf{NS}}  & {\textbf{UG}} & {{\textbf{DG}}} 
		& { AS$\uparrow$} \\\hline 
		Our \textbf{PLGM} &&&&24.36\\
		+ \textbf{NS} &$\checkmark$ & &  &  27.53\\
		+ \textbf{UG} & &$\checkmark$ & & 40.88 \\
		+ (\textbf{UG} +\textbf{UG})/Our \textbf{PLRM} & & $\checkmark$ & $\checkmark$ & \bf 41.74 \\
		\Xhline{1.2pt}
	\end{tabular}
\end{table}

\begin{table}\footnotesize
	\vspace{-0.05cm}
	\caption{The impact of the iteration number of the random walk operation on our method.  }
	\label{table_A4}
	\renewcommand\arraystretch{1.05}	
	\centering	

	\begin{tabular}{l@{\hskip 0.30cm}|c@{\hskip 0.3cm}c@{\hskip 0.3cm}c@{\hskip 0.3cm}c@{\hskip 0.3cm}c}
		\Xhline{1.2pt}
		{Iteration number}&{1}  & {5} & {10} & {20} & {$\infty$} \\ 
		\hline 
		{AS $\uparrow$} & 37.29 & 38.49 &  39.71 & 40.11 & \bf 41.74 \\
		\Xhline{1.2pt}
	\end{tabular}
\end{table}

\section{Conclusions}
In this paper, we propose a novel self-supervised scene flow learning method in point clouds to produce pseudo labels via point matching and perform pseudo label refinement by encouraging the local consistency.
Comprehensive experiments show that our method achieves state-of-the-art performance among self-supervised learning methods.
Our self-supervised method even performs on par with some supervised learning approaches, although we do not need any ground truth flow for training.

\section{Acknowledgements} 
This research was conducted in collaboration with SenseTime. 
This work is supported by A*STAR through the Industry Alignment Fund - Industry Collaboration Projects Grant.
This work is also supported by the National Research Foundation, Singapore under its AI Singapore Programme (AISG Award No: AISG-RP-2018-003), and the MOE Tier-1 research grants: RG28/18~(S) and RG22/19~(S).

{\small
\bibliographystyle{ieee_fullname}
\bibliography{egbib}
}

\end{document}